%% file: main.tex
\documentclass{article}
\usepackage{graphicx} % Required for inserting images
\usepackage{caption}
\usepackage{amsmath, amsfonts}
\usepackage{subcaption}

\usepackage[normalem]{ulem}
\usepackage{authblk}
\usepackage[utf8]{inputenc} 
\usepackage[T1]{fontenc}
\usepackage{lmodern}
\usepackage[hidelinks]{hyperref}

\usepackage{booktabs}
\usepackage{comment}
\usepackage{arxiv}
\usepackage{color}
\usepackage{soul}
\usepackage{fancyhdr}
\newcommand{\MMLU}{\mathrm{MMLU}}
\newcommand{\KL}{\mathrm{KL}}

\newbox{\myorcidaffilbox}

\sbox{\myorcidaffilbox}{\large\includegraphics[scale=0.7]{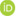}}
\newcommand{\orcidaffil}[1]{%
	\href{https://orcid.org/#1}{\usebox{\myorcidaffilbox}}}

\title{Full-Parameter Continual Pretraining of Gemma2: Insights into Fluency and Domain Knowledge}

\author{Vytenis \v{S}liogeris\orcidaffil{0000-0002-0943-9357}, Povilas Daniu\v{s}is\orcidaffil{0000-0001-5977-827X}, and Artūras Nakvosas\orcidaffil{0009-0007-7391-5454}}

\affil{Neurotechnology, Laisvės pr. 125A, LT-06118, Vilnius, Lithuania\thanks{Neurotechnology \url{http://www.neurotechnology.com} is a Lithuanian company, specialising in artificial intelligence, biometrics, computer vision, and deep neural networks.}}

\date{May 2025}

\begin{document}

\maketitle

\begin{abstract}
In this technical report, we empirically investigate the relationship between linguistic fluency and domain knowledge in the context of continual learning with large language models (LLMs). Specifically, we enhance the linguistic fluency of the Gemma2 LLM for the Lithuanian language by autoregressively pretraining its full parameter set on the first 10\% of the Lithuanian language component of the CulturaX dataset. To prevent catastrophic forgetting of the model's existing domain knowledge, we apply Elastic Weight Consolidation (EWC), leveraging Fisher information estimated using data from the Massive Multitask Language Understanding (MMLU) benchmark. In the post-training evaluations, we assess linguistic fluency through perplexity and evaluate domain knowledge using accuracy on a suite of language understanding benchmarks, including ARC-Easy, Belebele, GSM8K, HellaSwag, MMLU, TruthfulQA, and Winogrande, in both English and Lithuanian. The empirical results demonstrate that EWC not only mitigates catastrophic forgetting by preserving the model's performance in terms of both linguistic fluency and domain knowledge but also improves or maintains these capabilities for the newly added Lithuanian language. These findings highlight the potential for more efficient adaptation of general-purpose LLMs to under-represented languages without requiring access to the original training data. The accompanying codebase is openly accessible at: \url{https://github.com/Neurotechnology/LLM_EWC}.
\end{abstract}


\begin{comment}

% Perhaps important thoughts on the mathematics/relevance

The issue that prompted us to begin these investigations
is the catastrophic forgetting of mathematics when fine-tuning LLMs
on non-specialised datasets, such as CulturaX.
It was found that the model's accuracy at the gsm8k benchmark fell early in the training process
without any later improvement.
Mathematical reasoning is
a) semi-language agnostic
(mathematical symbols and equations transcend language, but proofs and reasoning are often accompanied by language) 
and b) quite specialised
(In the sense that  there is no hope of teaching the model mathematical reasoning without providing it with mathematical data.)

\end{comment}

\input{tex/introduction}

\input{tex/related_work}

\input{tex/experimental_setup}

\input{tex/results}

\input{tex/conclusions}

\bibliography{references}
%\printbibliography

\end{document}

%% file: tex/introduction.tex
\section{Introduction}

% Basic terminology
%\marginpar[Research question]{Basic terminology}

LLMs, based on Transformers~\cite{NIPS2017_3f5ee243} and other neural architectures, are remarkably efficient knowledge models. However, as with most neural networks, learning a new task often reduces the performance of the model on the previously learned tasks. This undesirable effect is known as \emph{catastrophic forgetting}. The field of mitigating catastrophic forgetting is called continual learning (CL). Specifically, it is a study of algorithms and models that can cope with a learning scenario in which new tasks should be learned, without losing performance on previously learned tasks. Formally, 
a model is usually represented by some parametrised function $f_{\theta}$ (e.g., neural network), with parameters $\theta$. Given a sequence of tasks $\mathcal{T}_{1},...,\mathcal{T}_{n}$ represented with data sets $\mathcal{D}_{1},...,\mathcal{D}_{n}$, which arrive over time, this model should be able to learn a new task $\mathcal{T}_{i}$ from $\mathcal{D}_{i}$, without access to previous $\mathcal{D}_{j}$ ($j<i$), simultaneously maintaining performance on all previously learned tasks $\mathcal{T}_{j}$ (where $j<i$). 

% Raison d'etre
%\marginpar[Research question]{Raison d'etre}

The intersection of CL and LLMs is a nascent field,
giving rise to questions regarding different types of knowledge.
Several knowledge ontologies have been proposed.
For example, a survey~\cite{wang2025bringknowledgesurveymethods} outlines factual knowledge, domain knowledge, language knowledge, and preference knowledge, and~\cite{shi2024continual} discusses different ontologies of knowledge,
pairing them with different CL methods to improve the particular type of knowledge.
In this work, we outline two types of knowledge: the knowledge of language fluency and domain knowledge.
Language fluency denotes the ability to produce grammatically correct sentences in a particular language. Computationally, it can be partially interpreted and measured via perplexity, which reflects how well the model can predict the next token, given the previous ones (lower perplexity indicates better performance).
Domain knowledge, on the other hand, describes the ability of the model to know and reason about a specific domain. Having a set of language understanding benchmarks, we can also evaluate the model's domain knowledge by investigating the accuracy of the answers it selects. These two types of knowledge are partially related to linguistic competence and linguistic performance~\cite{chomsky1965}.

%These types of knowledge are orthogonal to an extent;
%a model can clearly have good language fluency with no domain knowledge,
%and we can think of computer code as agents which have mathematical domain knowledge with no language fluency.
%We are interested in the question of whether or not improving linguistic fluency in a different language can improve domain knowledge expressed in that language.

%\marginpar[Research question]{Research question}

We are interested in the research question of whether an enhancement of linguistic fluency of a given language in LLM can also improve its domain knowledge in that language, simultaneously preserving LLM's linguistic fluency and domain knowledge in previously learned languages. Since it is a very general question, we provide only a partial analysis, focusing on Lithuanian and English, respectively, and using CL (in the form of EWC regularisation~\cite{kirkpatrick2017overcoming}) for the preservation of existing domain knowledge. 

Our main contributions to the posed research question consist of the empirical findings that when we enhanced Gemma2's Lithuanian language fluency via autoregressive pretraining using the first 10\% of the Lithuanian language component of the CulturaX dataset,

\begin{itemize}

\item EWC allowed us to mitigate the catastrophic forgetting effects in the English component both in linguistic fluency (measured via perplexity benchmark) and domain knowledge (measured via language understanding benchmarks; $7/7$ cases);

\item EWC allowed us to improve the performance of the Lithuanian component both in linguistic fluency (measured via perplexity benchmark) and domain knowledge in $5/7$ language understanding benchmarks (ARC-Easy, GSM8K, HellaSwag, MMLU, and Winogrande).
\end{itemize}

We structure our article beginning with a short review of the related work in Section~\ref{sec:related_work}. In Section~\ref{sec:experimental_setup} we discuss our experiment setup and present our empirical findings in Section~\ref{sec:results}. Finally, we conclude our research in Section~\ref{sec:conclusions}.

%% file: tex/related_work.tex
\section{Related work}
\label{sec:related_work}

\input{tex/related_work/continual_pretraining_on_llms}
\input{tex/related_work/elastic_weight_consolidation}

%% file: tex/related_work/continual_pretraining_on_llms.tex
\subsection{Continual learning in LLMs}

CL methods used in LLMs are roughly classified into replay-based, regularisation-based,  architecture-based~\cite{shi2024continual}, or combined approaches. Since in this research, we assume a two-task CL scenario, let us denote them with $\mathcal{A}$ and $\mathcal{B}$.

% Replay

Replay-based methods rely on a fraction of the previous task datasets to be preserved for future training, \cite{zheng2024breaking, scialom2022fine, ibrahim2024simple}.
However, previous training data are often not available, which limits the applicability of these methods for non-generative architectures. For generative architectures, the replay buffers can be sampled from the distribution, represented by the model, as in \cite{sun2019lamol}.

% Regularistaion
Regularisation-based methods rely on additive regularisers, which encourage CL. Having a model, trained for task $\mathcal{A}$, and new task $\mathcal{B}$, regularisation-based models minimise the regularised loss $L_{\mathcal{B}}(\theta) + \lambda \Omega_{\mathcal{A}}(\theta)$, where $\theta$ are the model's parameters, $L_{\mathcal{B}}(\theta)$ is loss for task $\mathcal{B}$, $\lambda > 0$ is a regularisation strength, and $\Omega_{\mathcal{A}}(\theta)$ is regulariser, which encourages CL. These methods are usually computationally efficient and can be combined additively. For instance, in EWC, $\Omega_{\mathcal{A}}(\theta) = \sum_{i} F_{i}(\theta_{i} - \theta_{\mathcal{A},i})^{2}$, where $F_{i}$ is Fisher information, $\theta_{i}$ is the $i$-th model's parameter, and $\theta_{\mathcal{A},i}$ is the corresponding parameter from the previous task (see Section~\ref{sec:ewc} for details). In synaptic intelligence $\Omega_{\mathcal{A}}(\theta) = \sum_{i}S_i(\theta_i - \theta_{\mathcal{A},i})^2$, where
$S_i = \frac{\sum_{t \leq T}(\theta_{\mathcal{A},i}(t+1) - \theta_{\mathcal{A},i}(t))\frac{\partial\mathcal{L}(t)}{\partial\theta_i[t]}}{(\theta_{\mathcal{A},i} - \theta_{\mathcal{A},i}(0) )^2 + \epsilon}$, $\epsilon > 0$ is damping parameter, and $t=0,1,...,T$ are training steps. This approach requires recording loss gradients and weight changes. In Learning without Forgetting~\cite{10.1007/978-3-319-46493-0_37}, 
$\Omega_{\mathcal{A}}(\theta) = \KL\left(f_{\mathcal{A}}(x)\|f_{\theta}(x)\right)$, where $f_A(x)$ is the previously trained model’s output distribution, $f_{\theta}(x)$ is the current model’s output distribution, and $\KL$ is the Kullback-Leibler divergence. This approach was used by~\cite{castellucci2021learning}
to mitigate catastrophic forgetting for semantic processing tasks. In their work, the BERT-based model is incrementally trained with new languages, using unlabeled data from previous tasks.

% <- tokiu detaliu, kurios neperteikia aiskaus vaizdo gal geriau nemineti

%This data is labelled with the teacher model, to use for training in conjunction with current task data.
%The requirement of unlabelled previous task data is too restrictive for contemporary LLMs
%as pretraining data is usually not available.

%A modified version of EWC has been implemented \cite{chen2020recall}.
%A combination of EWC and LoRA has also been showed to work for LLMs \cite{xiang2023language}.
%LoRA reduces the number of learnable parameters, which allows for training using less memory.

% Architecture-based
Architecture-based approaches aim to have specific architectural components for individual tasks. For example, different tasks can be learned using performance-efficient fine-tuning (PEFT) adapters, such as LoRA~\cite{hu2022lora} and CURLoRA~\cite{https://doi.org/10.5281/zenodo.12730055}.

Combined approaches integrate different CL methods to use the advantages and mitigate limitations of individual components (e.g., regularising PEFT adapters via EWC, as in~\cite{xiang2023language}). In addition, CL can be improved by incorporating various heuristics, such as learning rate schedulers~\cite{winata2023overcoming}, where the learning rate is reduced after each task as such: $lr_t = \max(lr_{min}, lr_{t-1} \cdot \gamma)$, where $\gamma > 0$ is a hyper-parameter. %Architecture-based and combined methods are considered among the most efficient (maybe true, but then why we use EWC). So removed this sentence. 

%% file: tex/related_work/elastic_weight_consolidation.tex
\subsection{Elastic Weight Consolidation}
\label{sec:ewc}

In our experiments, we will use EWC, motivated by recent findings that  regularisation-based CL methods, such as synaptic intelligence, are closely related to it~\cite{pmlr-v151-benzing22a}. The authors of~\cite{kirkpatrick2017overcoming} describe an EWC regularisation framework in which the parameters that are important for the previous tasks have reduced plasticity. 
In this framework, model optimization is the finding
of the most probable model parameters $\theta$ given data $\mathcal{D}$.
Taking logarithms of the Bayes' formula $p(\theta|\mathcal{D}) = \frac{p(\mathcal{D} | \theta)p(\theta)}{p(\mathcal{D})}$ we have:

\begin{equation}
    \log p(\theta|\mathcal{D}) = \log p(\mathcal{D} | \theta) + \log p(\theta) - \log p(\mathcal{D}),
\end{equation}

\noindent where $\log p(\mathcal{D}|\theta)$ describes the log-probability
of the model with parameters $\theta$ on the dataset $\mathcal{D}$,
which is the negative of the loss function.
In the CL scenario, where task $\mathcal{A}$ with data $\mathcal{D}_\mathcal{A}$ is followed by task $\mathcal{B}$ with data $\mathcal{D}_\mathcal{B}$,
the probability of model parameters $p(\theta)$ following a pretraining on dataset $\mathcal{D}_\mathcal{A}$
has value $p(\theta|\mathcal{D}_\mathcal{A})$.
Thus, the equation can be rewritten as follows:

\begin{equation}
    \log p(\theta|\mathcal{D}_\mathcal{B}) = \log p(\mathcal{D}_\mathcal{B} | \theta) + \log p(\theta|\mathcal{D}_\mathcal{A}) - \log p(\mathcal{D}_\mathcal{B}).
\end{equation}

We can notice that the right side of the equation depends on the loss of task $\mathcal{B}$,
and therefore all the information for task $\mathcal{A}$ is in the distribution $p(\theta|\mathcal{D}_\mathcal{A})$.
Transforming this into a loss function, we get eq. \ref{eq:EWC_loss}:

\begin{equation}\label{eq:EWC_loss}
    \mathcal{L}(\theta) = \mathcal{L}_{\mathcal{B}}(\theta) + \frac{\lambda}{2} \sum_{i}F_i(\theta_i-\theta_{A,i})^2,
\end{equation}

where $F_i = \mathbb{E} \left[ \left( \frac{\partial \log p(y | x; \theta_{\mathcal{A}})}{\partial \theta_i} \right)^2 \right] $ is Fisher information, and $p(y | x; \theta_{\mathcal{A}})$ is the conditional density. Intuitively, this regulariser measures the importance of the parameter $\theta_{i}$, and during the training of task $\mathcal{B}$ it penalises its deviation according to this importance.

To protect the parameters of Gemma2, which are potentially responsible for domain knowledge, we use MMLU data (which consists of a set of academic language understanding benchmarks in multiple domains) for Fisher's information estimation, estimating it via the empirical Fisher estimator (see Section 5.4 from~\cite{ven2025onthecomputation}):

\begin{equation}
F_{i} = \frac{1}{|D_{\MMLU}|}\sum_{(x,y) \in D_{\MMLU} } \left(  \frac{\partial}{\partial \theta_{i}} \log p_{\theta}(y|x) \bigg|_{\theta_{i}=\theta_{\mathcal{A},i}} \right)^2,
\end{equation}
where $D_{\MMLU}$ is the MMLU data set.

%% file: tex/experimental_setup.tex
\section{Experimental setup}
\label{sec:experimental_setup}

%For instance, using L\cite{li2024examining} probably do what we did??? }
%They do changing language.
%They use adapters and freezing parameters to append chinese language.
%One problem I see is that the tokens for chinese/english are very different,
%making the task very easy.
%Also their results seem too good to be true.

Conceptually, our experiment is similar to that of~\cite{li2024examining}, where the authors use various architecture-based methods to continually pretrain the Llama2 model for the Chinese language. However, instead of architecture-based methods, we focus on adding EWC regularisation for achieving CL.
In our experiments, we use Gemma2 LLM (gemma2-2b-it)~\cite{team2024gemma}, due to its sufficiently compact parametrisation and good performance on modern LLM benchmarks. 
The initial task $\mathcal{A}$ is the model's pretraining performed by the original authors, and task $\mathcal{B}$ is the next-token prediction on 10\% of the Lithuanian portion of CulturaX~\cite{nguyen2023culturax}.
We trained task $\mathcal{B}$ using EWC regularisation with a range of regularisation strengths $\lambda \in \{0, 10^2, 10^3, 10^6, 10^9, 10^{12} \}$, and evaluated the linguistic fluency and domain knowledge of the resulting models in English and Lithuanian. In the training process, we used the AdamW optimizer with cross-entropy loss and the following hyperparameters: a learning rate of $0.0002$, a warm-up ratio of $0.05$, weight decay of $0.01$, a per-device batch size of $2$, and $1$ gradient accumulation step. The evaluation of each of the $6$ values of $\lambda$ described above required approximately $4$ hours, totalling $24$ hours of computation time. All experiments were carried out on a cluster of $8$ H100 GPUs.

\textbf{Linguistic fluency.} For this, we performed two perplexity benchmarks. The first one was aimed at assessing the effectiveness of the CL in terms of perplexity. Specifically, we evaluated the average perplexity of TruthfulQA question-answer pairs (in both English and Lithuanian). In the second one, we investigated the potential negative effects of EWC for the added Lithuanian language, when the regularisation strength $\lambda$ is excessively high. To this end, we used the Lithuanian Q/A dataset~\cite{nakvosas2024open, nakvosas2024open_infor} and measured the average perplexity of the model's responses to questions from this dataset using LT-Llama2-13B, which is noted for its grammatical accuracy~\cite{dzikiene2025localizingaievaluatingopenweight}.  

\textbf{Domain knowledge.} To measure the model's domain knowledge, we used popular language understanding benchmarks listed in Table~\ref{tab:llm-benchmarks} (both in English and Lithuanian~\cite{nakvosas2024open, nakvosas2024open_infor}). Note that although the English version of MMLU data was used in EWC to estimate Fisher information, we included this dataset in our benchmarks because, in our opinion, its empirical performance still may be interesting.

%lr: 2e-4
%warmup_ratio: 0.05
%weight_decay: 0.01
%Per-device batch size: 2
%Gradient accumulation steps: 1

\begin{table}[htbp]
\centering
%\resizebox{\textwidth}{!}{%
\begin{tabular}{|l|l|l|l|}
%\toprule
\hline
\textbf{Benchmark} & \textbf{Description} & \textbf{\# Instances} & \textbf{Reference(s)} \\
\midrule
\hline
MMLU & Multi-task language understanding across 57 tasks. & 15,908 & 
\cite{hendrycks2021ethics,hendryckstest2021} \\
Belebele & Multilingual reading comprehension benchmark. & 122,000 & \cite{bandarkar-etal-2024-belebele} \\
GSM8K & School-level math word problems. & 8,500 & \cite{cobbe2021gsm8k} \\
HellaSwag & Commonsense reasoning completion tasks. & 70,000 & \cite{zellers2019hellaswag} \\
ARC-Easy & School-level science questions (easy subset). & 2,251 & \cite{allenai:arc} \\
TruthfulQA & Assessing truthfulness of model-generated answers. & 817 & \cite{lin2021truthfulqa} \\
WinoGrande & Commonsense reasoning with pronoun resolution. & 44,000 & \cite{ai2winogrande} \\
\hline
%\bottomrule
\end{tabular}
\caption{Summary of performed language understanding benchmarks.}
\label{tab:llm-benchmarks}
\end{table}

%% file: tex/results.tex
\section{Results}
\label{sec:results}

\begin{figure}[ht]
    \centering
    \begin{subfigure}[t]{0.49\textwidth}
        \centering
        \includegraphics[width=\textwidth]{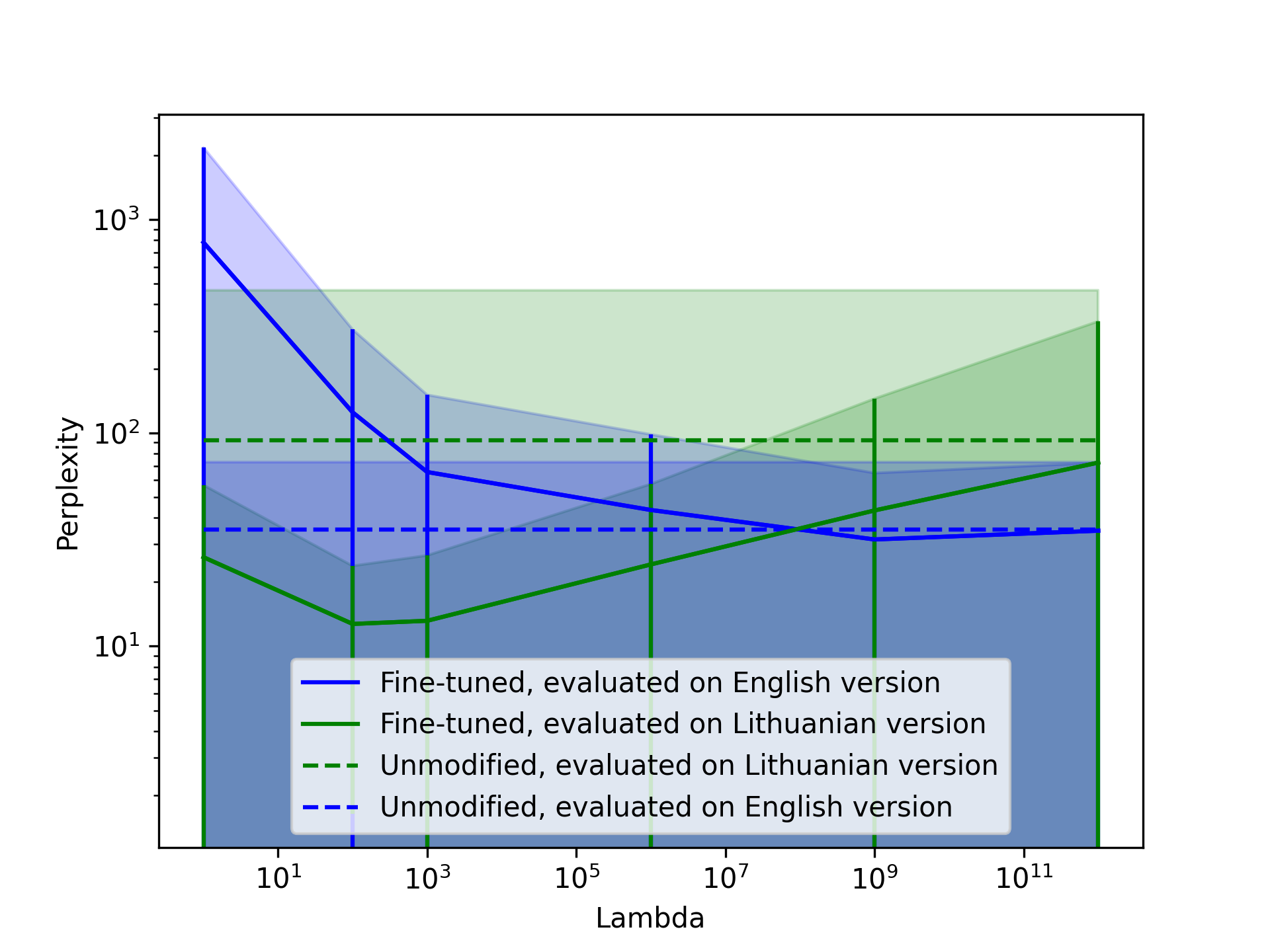}
        \caption{Perplexity vs. regularisation strength $\lambda$ (TruthfulQA data).}
        \label{fig:perplexity}
    \end{subfigure}    
    \hfill
    \begin{subfigure}[t]{0.49\textwidth}
        \centering
        \includegraphics[width=\textwidth]{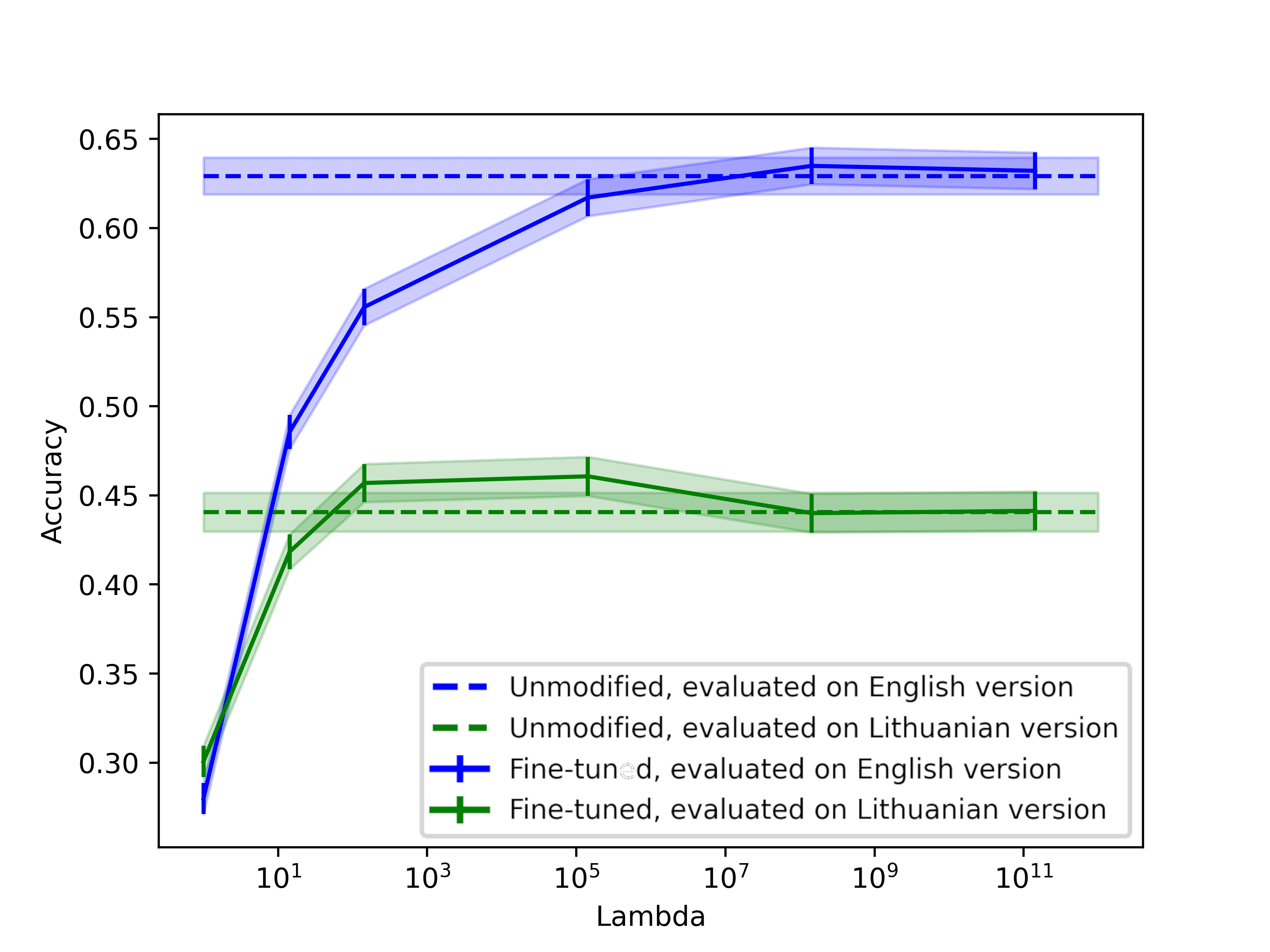}
        \caption{The averaged accuracy on all language understanding benchmarks versus regularisation strength $\lambda$.}
        \label{fig:average_acc}
    \end{subfigure}
    \caption{Comparison of perplexity and average accuracy in domain understanding tasks with varying regularisation strength $\lambda$. $\lambda = 0$ denotes a setting without EWC. "Fine-tuned" plot indicates autoregressive pretraining with EWC regularisation.}
    \label{fig:comparison}
\end{figure}

\begin{figure}[b]
    \centering
    \includegraphics[scale=0.12]{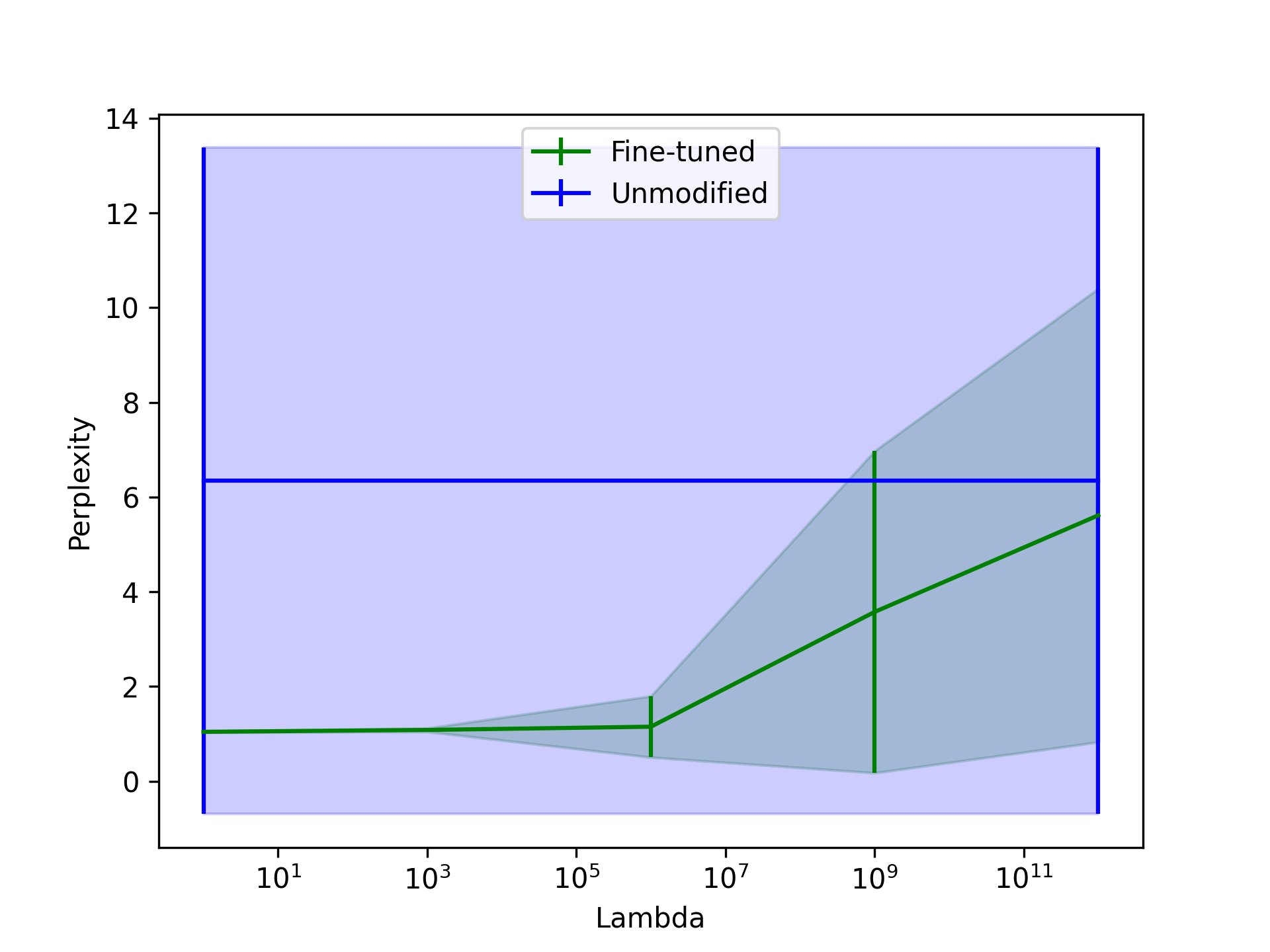}
    \caption{Perplexity of the answer (evaluated with LT-Llama-13B) versus regularisation strength $\lambda$ (Lithuanian Q/A~\cite{nakvosas2024open,nakvosas2024open_infor} data).}
    \label{fig:lt_ppl}
\end{figure}

\textbf{Linguistic fluency.} 
Figure~\ref{fig:perplexity} shows the average perplexity evaluated using the TruthfulQA data in the same manner, while Figure~\ref{fig:lt_ppl} displays the average perplexity evaluated via LT-Llama2-13B on the Lithuanian Q/A dataset. Figure~\ref{fig:perplexity} shows that the EWC enabled the preservation of English data perplexity when the Lithuanian language was integrated into the model. Although with $\lambda = 0$, we observed a similar effect as in domain knowledge benchmarks, as the value of $\lambda$ increases, the perplexity of the English data approaches that of the initial Gemma2 model. On the other hand, Figure~\ref{fig:lt_ppl} shows surprisingly low perplexities for $\lambda < 10^9$. As $\lambda$ increases, the perplexity of the answers tends to rise, indicating the negative effects of overly strong regularisation.

\begin{figure}[ht]
    \centering
    \includegraphics[scale=0.55]{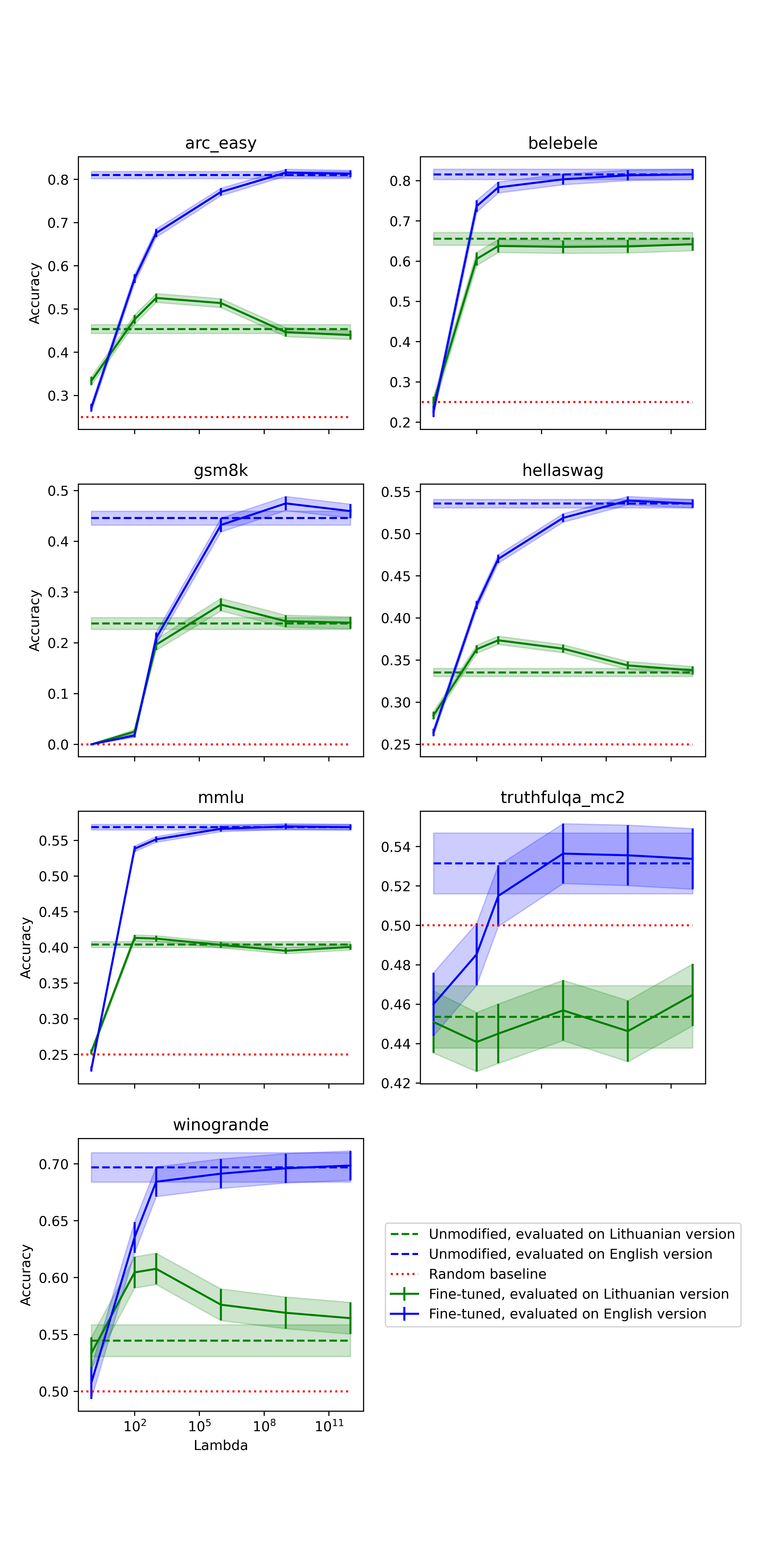}
    \caption{Accuracy of the models versus EWC regularisation strength $\lambda$ on the language understanding benchmarks. "Fine-tuned" plot indicates autoregressive pretraining with EWC regularisation.}
    \label{fig:lmeh_acc}
\end{figure}

\textbf{Domain knowledge.}
Figure~\ref{fig:average_acc} presents the average accuracy across all language understanding benchmarks listed in Table~\ref{tab:llm-benchmarks} for different values of $\lambda$. 
The results of the individual language understanding benchmarks can be found in Figure~\ref{fig:lmeh_acc}. It can be seen that with $\lambda=0$, analogous to not using EWC at all, the performance of the model drops significantly, often even not reaching the initial accuracy. This may be because our dataset (10\% of the Lithuanian component of CulturaX, which mainly consists of web crawls of common websites) was insufficient for an improvement of LLMs trained with much larger and more diverse data. On the other hand, $\lambda > 10^{11}$ describes the case where the model is non-plastic. This can be seen in the accuracy, which is very similar to that of the initial model, suggesting that the model likely did not change much from its initial version, trained on task $\mathcal{A}$. The range $\lambda \in [10^2, 10^{11}]$ reveals two interesting effects. First, in Figure~\ref{fig:lmeh_acc} we see that in this range the accuracies for the Lithuanian version of the benchmarks are also higher, indicating that EWC may be helpful not only to not forget domain-level knowledge in English but also to attain it more efficiently in the newly added Lithuanian language. This may be partially explained by the mechanism of EWC in our setup, which inhibits updates of the parameters that are important for domain knowledge. In addition, Figure~\ref{fig:lmeh_acc} shows that for larger $\lambda$, EWC regularisation may even increase the accuracy for English domain knowledge benchmarks (e.g., GSM8K, TruthfulQA sets).

Figure \ref{fig:lmeh_acc} includes the evaluation of the models on the GSM8K benchmark, which consists of grade-school mathematical problems (see Table~\ref{tab:llm-benchmarks}). It was previously observed~\cite{nakvosas2024open,nakvosas2024open_infor} that the Llama2 model, fine-tuned on the Lithuanian part of the CulturaX dataset,
loses its mathematical ability due to the absence of mathematics in CulturaX data.  Although in our experiment we used a different LLM architecture, this effect is also visible in Figure~\ref{fig:lmeh_acc}. This figure also shows that for $\lambda > 10^6$, the mathematical ability of Gemma2 LLM is retained with the help of EWC.

Figure~\ref{fig:comparison} suggests that the values of $\lambda$ that resulted in lower perplexity also correspond to better performance in domain knowledge benchmarks, partially in agreement with the findings of~\cite{gonen-etal-2023-demystifying}.

%% file: tex/conclusions.tex
\section{Conclusions}
\label{sec:conclusions}

% Italready knew some LT.
% When trained without EWC (lambda = 0) even not reaches performance of initial model.

We empirically investigated the posed research question of whether an enhancement of the linguistic fluency of the Lithuanian language in LLM can also improve its domain knowledge in that language, simultaneously preserving LLM's linguistic fluency and domain knowledge in previously learned English language. We used perplexity as a measure of linguistic fluency and evaluated the model's domain knowledge via the accuracies of popular language understanding benchmarks (ARC-Easy, Belebele, GSM8K, HellaSwag, MMLU, TruthfulQA, and Winogrande). In order to preserve the existing knowledge, we used CL in the form of EWC.

Specifically, we autoregressively pretrained the Gemma2 LLM (2 billion parameter version) with 10\% of the Lithuanian component of CulturaX, using EWC regularisation with different regularisation strengths $\lambda$. To foster reproducible research, we report the hyper-parameters we used in our experiments and include a link to the accompanying code repository.

The experiments performed reveal that our setup allowed us to enhance the Lithuanian component of Gemma2, simultaneously mitigating the catastrophic forgetting effects in its English component 
in both linguistic fluency and domain knowledge on all $7$ benchmarks (ARC-Easy, Belebele, GSM8K, HellaSwag, MMLU, TruthfulQA, and Winogrande). Furthermore, the EWC regularisation also improved both linguistic fluency and domain knowledge for the Lithuanian language on the $5/7$ benchmarks (ARC-Easy, GSM8K, HellaSwag, MMLU, and Winogrande).

These findings support an affirmative response to the research question posed. Our results may have practical implications for utilising general-purpose LLMs for specialisation in low-resource languages: if it were possible to shift linguistic fluency without disturbing the domain knowledge of an LLM, the creation of stronger regional LLMs would become far easier. 

\noindent \textbf{Limitations.} Although we conducted a fairly large experiment, our findings are still based on limited data. For example, we used only 10\% of the Lithuanian component of CulturaX for autoregressive pretraining, evaluated the Fisher information using MMLU only, and relied on just TruthfulQA and Lithuanian Q/A datasets for perplexity evaluations. Our approach to evaluating Fisher information via MMLU also asks for a theoretical justification. In addition, comparisons with other CL methods would better connect our work with the existing body of research and elucidate the extent to which our empirical findings hold. Although we roughly estimated intervals, the selection of optimal regularisation strength $\lambda$ when applying EWC to full parameter continual pretraining of LLMs is still an open question.

\noindent \textbf{Future work.} We plan to investigate combined CL approaches for generative LLMs, leveraging their ability to generate samples from the initial task distribution and exploring mechanisms that allow for sparse updates while compensating for the limitations of individual components. In addition, since the linguistic fluency and domain knowledge of language models can be measured through perplexity and accuracy in benchmarks, a causal investigation of these two signals, in our opinion, would potentially provide interesting insights about LLMs.

\section*{Acknowledgement}

This research was funded and conducted by Neurotechnology. We are grateful to Neurotechnology for providing the necessary resources and support. We also thank Greta Tikužytė for editing the English language, and our colleagues for their helpful comments and discussions.

%{\color{blue} In addition, our empirical findings motivate further investigation of the relationship between linguistic competence and linguistic performance~\cite{chomsky1965} in LLMs.}

\begin{comment}
Due to the peculiarities of the implementation, some common optimizations, such as parts of ZeRO \cite{rajbhandari2020zero}, may not be possible.
Furthermore, since loss requires model parameters on task $A$, the memory usage during training increases with the size of the model. Also, the diagonal Fisher information matrix, on which EWC relies, is the same size as the parameters of LLM.
\end{comment}